\documentclass[runningheads]{llncs}
\usepackage[T1]{fontenc}
\usepackage{graphicx}
\usepackage[caption=false]{subfig}
\usepackage{booktabs}
\usepackage{multirow}
\usepackage{amsmath}
\begin{document}
\title{Learning to Segment using Summary Statistics and Weak Supervision}

\author{Omkar Kulkarni\inst{1}\orcidID{0009-0007-1405-6442} \and
Edward Raff\inst{2}\orcidID{0000-0002-9900-1972} \and
Tim Oates\inst{1}\orcidID{0000-0002-8655-747X}}
\authorrunning{Omkar Kulkarni et al.}
%
\institute{University of Maryland, Baltimore County, Baltimore MD, USA
\email{\{omkar.kulkarni,oates\}@umbc.edu}
\and
Crowdstrike, Austin TX, USA
\email{edward.raff@crowdstrike.com} 
}
\maketitle              
\begin{abstract}
Medical experts often manually segment images to obtain diagnostic statistics and discard the resulting annotations.
We aim to train segmentation models to alleviate this burden, but constrained to the retained summary statistics (e.g., the area of the annotated region).  Empirical results suggest that statistics alone are insufficient for this task, but adding weak information in the form of a few pixels within the area of interest significantly improves performance. 
We use a novel loss function that combines terms for image reconstruction quality, matching to summary statistics, and overlap between the predicted foreground and the weak supervisory signal.  Experiments on standard image, ultrasound (breast cancer), and Computed Tomography (CT) scan (kidney tumors) data demonstrate the utility and potential of the approach.

\end{abstract}

\section{Introduction}\label{sec:introduction}

\begin{figure}[t]
    \centering
    \subfloat[Input Image]{
        \includegraphics[width=0.16\columnwidth]{Example_Images/BUSI/OG.png}
    } 
    \subfloat[Ground \\truth]{
        \includegraphics[width=0.16\columnwidth]{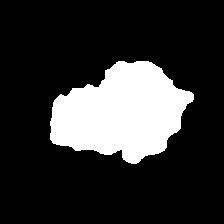}
    } 
    \subfloat[Weak mask]{
        \includegraphics[width=0.16\columnwidth]{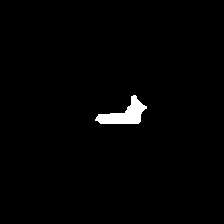}
    } 
    \subfloat[Our \\Method]{
        \includegraphics[width=0.16\columnwidth]{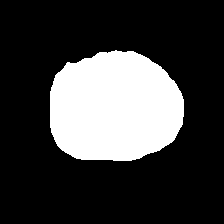}
    }
    \subfloat[Only weak mask]{
        \includegraphics[width=0.16\columnwidth]{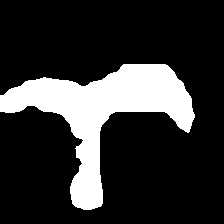}
    }
    \subfloat[Only statistics]{
        \includegraphics[width=0.16\columnwidth]{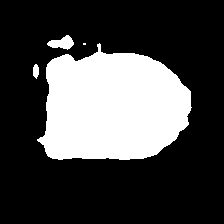}
    }
    \caption{
        Example images from the BUSI dataset, showing that we can obtain reasonably accurate segmentations even under highly restrictive and minimal learning signals from the physician.
    }  
    \label{fig:busi_example}
\end{figure}

Our work is motivated by an attempted collaboration with physicians at the University of Maryland School of Medicine. Multiple research projects, such as \cite{Klein2019,KOLAKOWSKI20181539,valencia2018}, involved months of manual effort by physicians and physician trainees 
to label images pixel-by-pixel and categorize cell types. The ultimate goal of this activity was to produce a clinically relevant ratio for the subject under study. Our interest in helping these physicians accelerate their work by using AI to segment the images was thwarted by the discovery that the original segmentation masks, the ground truth labels we would train and evaluate on, were routinely discarded as ``not relevant'' to the clinical question. \textit{Only the ratio was saved}. The effort to relabel even a few hundred images was measured in half-year increments, depending on the project and the required details. This problem motivates our work to develop a new approach to semi-supervised segmentation that directly addresses this common clinical scenario. Despite the difficulty of learning under these constraints, our results can be surprisingly useful --- as demonstrated in Figure \ref{fig:busi_example}.

Numerous prior works have investigated medical image segmentation as summarized in \cite{RAYED2024101504}, however, these are all fully supervised methods. Recent approaches toward weakly/semi supervised segmentation use skeletonization methods to encode domain knowledge, regularization algorithms, or incomplete labels (most similar to our work) \cite{das_miccai2024}. To wit, none have dealt with the kind of label scarcity or interdisciplinary hurdles to leveraging historical data. In tackling this problem, we find that a combination of self-supervised and weak signals can deliver practical and reasonable results. 

\section{Methods and Architecture} \label{sec:method}

We use a \textbf{percentage of ROI (Region of Interest)} as a summary statistic that represents the ROI's size relative to the image's size. Physicians conducting research often record this information for analysis, e.g., the percentage of fat cells vs. regular cells in \cite{valencia2018}. Physicians calculate this by labeling the image, and then the labeled image is discarded for perceived uselessness, recording only the summary statistic for posterity. It is simply the fraction of pixels in the ROI to the total number of pixels in the image. Mathematically, this is defined as $\bar{M} = \frac{1}{I \times J} \sum_{i=1}^{I} \sum_{j=1}^{J} M_{i,j}$.
As noted, our experiments use a ratio equal to the ground-truth fraction of pixels in the segmentable class. A degenerate solution we encountered is that the sigmoid prediction mask would learn to produce a constant value near the target ratio. While this got the correct ``summary statistic'', it failed to meaningfully segment any of the image content. 

To tackle this, we came up with a \textbf{`weak mask'} that contains information about the location and shape of the ROI. Let $X$ be the medical image, $M$ be its corresponding ground truth mask, and $M_w$ be the weak mask. It is easier to obtain a coarse localization of the ROI, typically at its center, than to accurately hand-annotate and obtain $M$. Thus for a given $X$, it is easier to obtain $M_w$ than to obtain $M$, i.e., $M_w \subset M$. Since $M$ already exists for the datasets used in this research, we obtain $M_w$ from $M$ by eroding $M$ to mimic how a radiology technician would obtain $M_w$ from $X$ - annotate only some pixels of the ROI, primarily the ones at the center.

We also want to ensure that the predictions are highly confident, so we avoid output values close to 0.5 or any other constant ratio. To achieve this, we implement a ``confidence'' loss function. We initialize a constant matrix $I_c$ with each element equal to 0.5 and with the same size as $M_p$. If $M_p$ is the predicted mask, the confidence loss here is defined as $L_c = \frac{1}{\operatorname{size}(I_c)}\sum_{i,j}(I_c)_{i,j}^2- \frac{1}{\operatorname{size}(I_c)}\sum_{i, j}(I_c - M_p)_{i,j}^2$.
Thus, each element that is 0.5 accrues the maximum penalty possible of 0.25. This encourages the model to output near-0 or near-1 values across the matrix and avoids degenerate solutions. 
The second form of self-supervision is a \textbf{reconstruction loss}, $L_r(I, Y) = \|I - Y\|_1$, as in an autoencoder. 
If the network can use the intermediate representation to reconstruct the input, it has likely learned to encode structural patterns in the input image that can improve segmentation. Since the loss $L_r$ and the preceding confidence loss $L_c$ are used in all experiments, they will not be stated explicitly when their inclusion would clutter the page or diagram. 
We now describe the \textbf{weak supervision loss ($L_{ws}$)} between the predicted mask and the weak mask as follows: Let $M_p$ be the predicted mask, $M$ be the ground truth mask, and $M_w$ be the weak mask. We only have the weak mask and assume no knowledge of the remaining pixels. Hence, we consider only the pixels for which we know information when computing a loss. We do an element-wise multiplication of the weak mask with the predicted mask, thus zeroing out all other pixels. The loss is then $L_{ws}(M_w, M_p) = -[M_w \odot\log(M_w \odot M_p) + (1 - M_w) \log(1 - M_w \odot M_p)]$, where `$\odot$' denotes element-wise multiplication. By enforcing that the weak mask is sufficiently small for a domain expert to quickly mark the unambiguous portion of an image (i.e., `center mass' avoiding the edges), we replicate and evaluate how much expert physician relabeling may be needed in future studies. 
The \textbf{statistics loss ($L_s$)} we use is simply the L1 loss between the percentage of ROI
in the predicted mask and the actual mask, $L_s(M,M_p) = \|\bar{M} - \bar{M}_p\|_1$. 

\section{Results} \label{sec:results}
We demonstrate our method on two datasets. 
The Breast Ultrasound Tumor (BUSI) dataset \cite{busi} and the KiTS23 dataset \cite{kits}. Volumetric data are treated as 2D in our tests by selecting the volume slice with the largest ROI. We use the DeepLabV3 \cite{deeplabv3} architecture with a ResNet-50 \cite{resnet50} backbone across all our experiments. In addition to the primary predictions from segmentation maps, we use the auxiliary output as an autoencoder, forcing it to reconstruct the input image. This helps preserve the image's 2D structure, which is useful for segmentation under weak supervision. 
We use IoU (Intersection over Union) as our standard evaluation metric.
For both medical imagery datasets, the best performance is seen in the setting where both losses, $L_s$ and $L_{ws}$, are used. The summary statistics loss ($L_s$) appears to be more significant than the weak supervision loss ($L_{ws}$). Table \ref{tab:results} summarizes our results from each dataset in all the settings tested.
\begin{table}[t]
\centering
\small 
\caption{IoU achieved with three weak mask sizes ([4-12\%]), ablating where either one of the losses is disabled, as well as with a fully supervised scenario. The table shows that we can achieve results that are not far from those of fully supervised learning by combining the $L_s$ and $L_{ws}$ loss functions. Also to be noted is that both loss functions are necessary to achieve this result.
}
\setlength{\tabcolsep}{4pt} 
\begin{tabular}{|l |c| ccc |c| c|}
\toprule
Dataset & $L_s$ only & 4\% & 8\% & 12\% & $L_{ws}$ only & Fully Supervised \\ 
\midrule
BUSI    & 36.95          & 45.70 & \textbf{48.13} & 45.29 & 26.33 & 63.15 \\
KiTS23  & 12.65          & 29.22 & \textbf{37.23} & 28.43 & 8.75  & 59.85 \\ 
\bottomrule
\end{tabular}
\label{tab:results}
\end{table}
\begin{figure}[t]
    \centering

    \subfloat[Input CT scan]{\includegraphics[width=0.16\columnwidth]{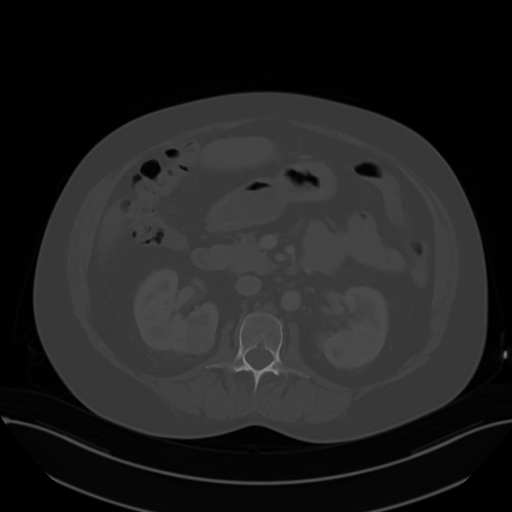}}\hfill
    \subfloat[Kidney mask]{\includegraphics[width=0.16\columnwidth]{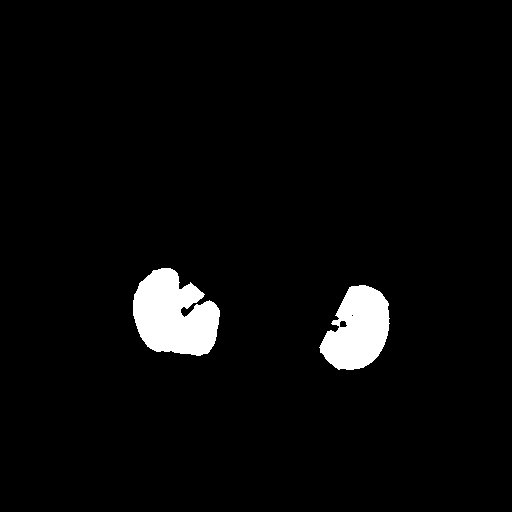}}\hfill
    \subfloat[Tumor mask]{\includegraphics[width=0.16\columnwidth]{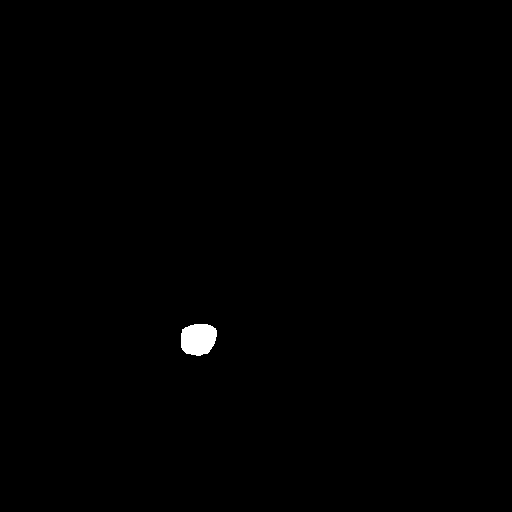}}\hfill
    \subfloat[Weak kidney mask]{\includegraphics[width=0.16\columnwidth]{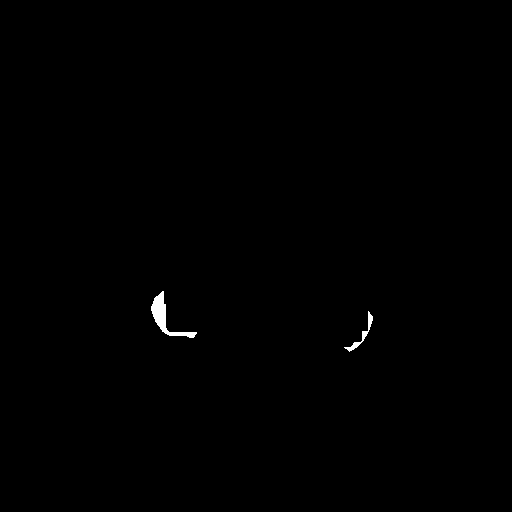}}\hfill
    \subfloat[Weak tumor mask]{\includegraphics[width=0.16\columnwidth]{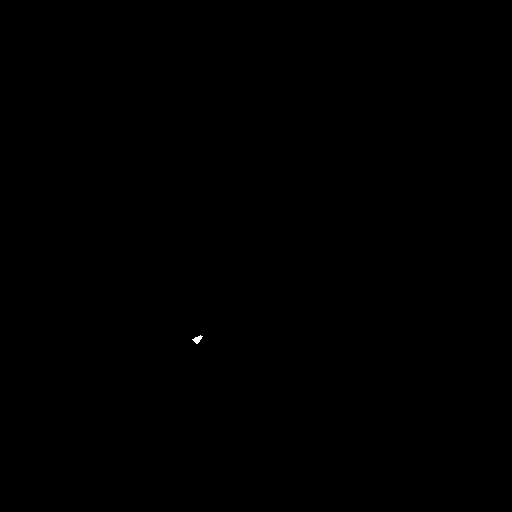}}


    \subfloat[Kidney: $L_s$]{\includegraphics[width=0.16\columnwidth]{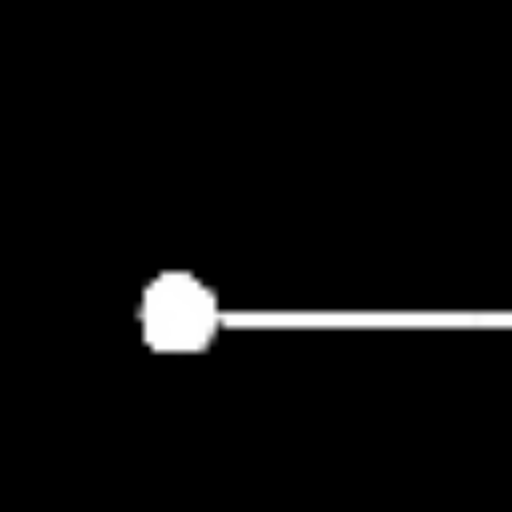}}\hfill
    \subfloat[Kidney: $L_{ws}$]{\includegraphics[width=0.16\columnwidth]{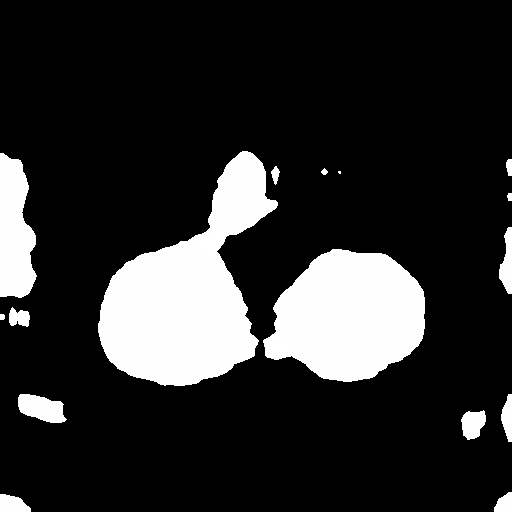}}\hfill
    \subfloat[Kidney: $L_s, L_{ws}$]{\includegraphics[width=0.16\columnwidth]{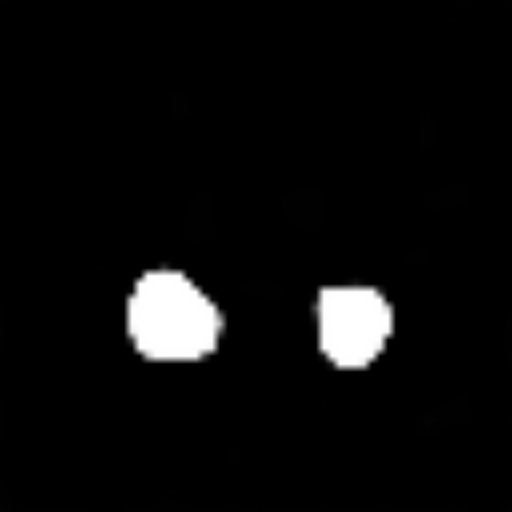}}\hfill
    \subfloat[Tumor: $L_s$]{\includegraphics[width=0.16\columnwidth]{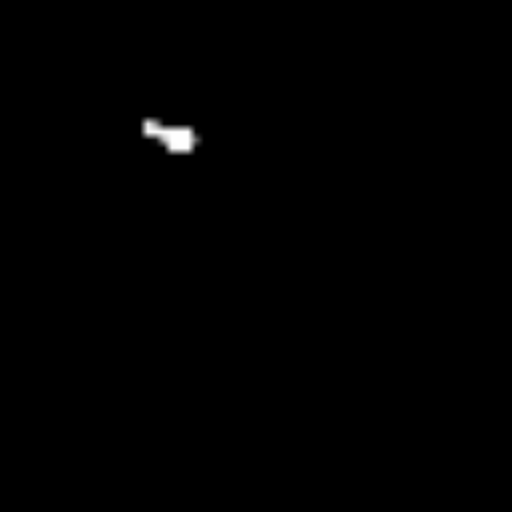}}\hfill
    \subfloat[Tumor: $L_{ws}$]{\includegraphics[width=0.16\columnwidth]{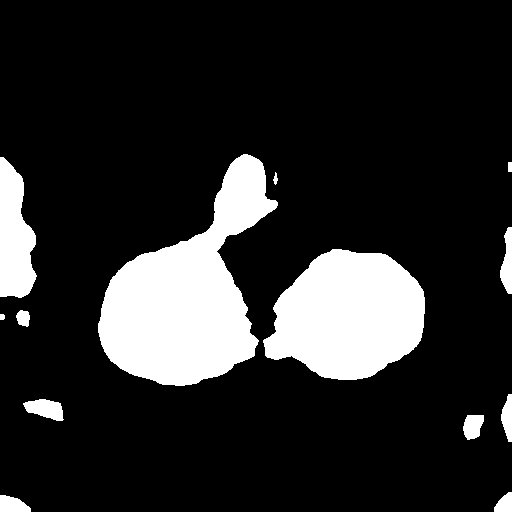}}\hfill
    \subfloat[Tumor: $L_s, L_{ws}$]{\includegraphics[width=0.16\columnwidth]{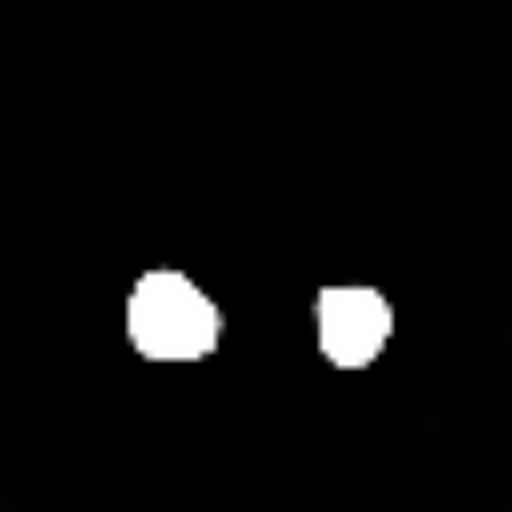}}

    \caption{A sample from the KiTS dataset, showing reasonably accurate segmentation masks even under highly restrictive and minimal learning signals. While the tumor prediction is less accurate, the kidney prediction is very close to the ground truth mask.}
    \label{fig:KiTS_example}
\end{figure}

An example image from the KiTS dataset is shown in Fig. \ref{fig:KiTS_example}, which illustrates the benefits and risks of our strategy. The top row shows the ground-truth and weak ground-truth used in training. The bottom row shows the kidney and tumor predictions for the three combinations of $L_s$ and $L_{ws}$.  Notice that the kidney mask is produced with high accuracy despite the limited labeling, whereas the tumor prediction has devolved to predicting the kidneys, resulting in a much lower 7\% IoU compared to 66\% for the kidney. 

At the macro scale, the image poses the problem that the mask is small over a non-visually distinct portion of the image for the tumor, which has a similar scale ratio of mask foreground to background pixels. This makes it difficult for the model to determine which actual content is intended to be segmented, especially since there is no negative signal in invalid regions to select (i.e., the weak segmentation loss $L_{ws}$ has only positive signals, with no negative signals). When combined with the reconstruction loss, this means the model can misattribute a segmentation signal to a different, more visually distinct, but similarly sized organ (in this case, the kidneys).

This result indicates that, in our current results, medical tasks with more visually distinct content will be better application areas. This is still a large time savings, as informed by our physician counterparts, especially when considering irregular cell structures that may have highly irregular borders but be visually distinct from the surrounding tissue. 

\section{Conclusion} \label{sec:conclusion}

Our results show that it is possible to obtain reasonably accurate segmentations with weak supervision without high-quality ground truth for medical image data. Our method performs best when the content of interest is visually distinct from the surrounding tissue in medical images.  We successfully demonstrate our method on ultrasound and CT scan imagery under a highly restrictive labeling scenario prevalent in hospitals today, proving that there is still value in the limited labels available. We plan to expand this to 3D data, and develop this further for probabilistic masking in the future.

\bibliographystyle{splncs04}
\bibliography{mybibliography}
\end{document}